\documentclass{article}

\usepackage{microtype}
\usepackage{graphicx}
\usepackage{subfigure}
\usepackage{booktabs} 
\usepackage{multirow} 

\usepackage{hyperref}

\usepackage[accepted]{icml2020}

\icmltitlerunning{Deep Nearest Neighbor Anomaly Detection}

\begin{document}

\twocolumn[
\icmltitle{Deep Nearest Neighbor Anomaly Detection}

\icmlsetsymbol{equal}{*}

\begin{icmlauthorlist}
\icmlauthor{Liron Bergman}{equal,huji}
\icmlauthor{Niv Cohen}{equal,huji}
\icmlauthor{Yedid Hoshen}{huji}
\end{icmlauthorlist}

\icmlaffiliation{huji}{School of Computer Science and Engineering, The Hebrew University of Jerusalem, Israel}

\icmlcorrespondingauthor{Yedid Hoshen}{yedid.hoshen@mail.huji.ac.il}

\icmlkeywords{Machine Learning, ICML}

\vskip 0.3in
]



\printAffiliationsAndNotice{\icmlEqualContribution} 

\begin{abstract}
Nearest neighbors is a successful and long-standing technique for anomaly detection. Significant progress has been recently achieved by self-supervised deep methods (e.g. RotNet). Self-supervised features however typically under-perform Imagenet pre-trained features. In this work, we investigate whether the recent progress can indeed outperform nearest-neighbor methods operating on an Imagenet pretrained feature space. The simple nearest-neighbor based-approach is experimentally shown to outperform self-supervised methods in: accuracy, few shot generalization, training time and noise robustness while making fewer assumptions on image distributions.
\end{abstract}
\section{Introduction}
\label{sec:intro}

Agents interacting with the world are constantly exposed to a continuous stream of data. Agents can benefit from classifying particular data as anomalous i.e. particularly interesting or unexpected. Such discrimination is helpful in allocating resources to the observations that require it. 
This mechanism is used by humans to discover opportunities or alert of dangers. Anomaly detection by artificial intelligence has many important applications such as fraud detection, cyber intrusion detection and predictive maintenance of critical industrial equipment. 

In machine learning, the task of anomaly detection consists of learning a classifier that can label a data point as normal or anomalous. In supervised classification, methods attempt to perform well on normal data whereas anomalous data is considered noise. The goal of an anomaly detection methods is to specifically detect extreme cases, which are highly variable and hard to predict. This makes the task of anomaly detection challenging (and often poorly specified). 

The three main settings for anomaly detection are: supervised, semi-supervised and unsupervised. In the \textit{supervised} setting, labelled training examples exist for normal and anomalous data. It is therefore not fundamentally different from other classification tasks. This setting is also too restrictive for many anomaly detection tasks as many anomalies of interest have never been seen before e.g. the emergence of new diseases. In the more interesting \textit{semi-supervised} setting, all training images are normal with no included anomalies. The task of learning a normal-anomaly classifier is now one-class classification. The most difficult setting is \textit{unsupervised} where an unlabelled training set of both normal and anomalous data exists. The typical assumption is that the proportion of anomalous data is significantly smaller than normal data. In this paper, we deal both with the semi-supervised and the unsupervised settings. 
Anomaly detection methods are typically based on distance, distribution or classification. The emergence of deep neural networks has brought significant improvements to each category. In the last two years, deep classification-based methods have significantly outperformed all other methods, mainly relying on the principle that classifiers that were trained to perform a certain task on normal data will perform this task well on unseen normal data, but will fail on anomalous data, due to poor generalization on a different data distribution. 

In a recent paper, \citet{gu2019statistical} demonstrated that a K nearest-neighbours (kNN) approach on the raw data is competitive with the state-of-the-art methods on tabular data. Surprisingly, kNN is not used or compared against in most current image anomaly detection papers. In this paper, we show that although kNN on raw image data does not perform well, it outperforms the state of the art when combined with a strong off-the-shelf generic feature extractor. Specifically, we embed every (train and test) image using an Imagenet-pretrained ResNet feature extractor. We compute the K nearest neighbor (KNN) distance between the embedding of each test image and the training set, and use a simple threshold-based criterion to determine if a datum is anomalous. 

We evaluate this baseline extensively, both on commonly used datasets as well as datasets that are quite different from Imagenet. We find that it has significant advantages over existing methods: i) higher than state-of-the-art accuracy ii) extremely low sample complexity iii) it can utilize very strong external feature extractors, at minimal cost iv) it makes few assumptions on the images e.g. images can be rotation invariant, and of arbitrary size v) it is robust to anomalies in the training set i.e. it can handle the unsupervised case (when coupled with our two-stage approach) vi) it is plug and play, does not have a training stage.

Another contribution of our paper is presenting a novel adaptation of kNN to image group anomaly detection, a task that received scant attention from the deep learning community. 

Although using kNN for anomaly detection is not a new method, it is not often used or compared against by most recent image anomaly detection works. Our aim is to bring awareness to this simple but highly effective and general image anomaly detection method. We believe that every new work should compare to this simple method due to its simplicity, robustness, low sample complexity and generality.

\section{Previous Work}
\label{sec:prev}

\textit{Pre-deep learning methods:} The two classical paradigms for anomaly detection are: reconstruction-based and distribution-based. Reconstruction-based methods use the training set to learn a set of basis functions, which represent the normal data in an effective way. At test time, they attempt to reconstruct a new sample using the learned basis functions. The method assumes that normal data will be reconstructed well, while anomalous data will not. By thresholding the reconstruction cost, the sample is classified as normal or anomalous. Choices of different basis functions include: sparse combinations of other samples (e.g. kNN) \citep{eskin2002geometric}, principal components \citep{jolliffe2011principal, candes2011robust}, K-means \citep{hartigan1979algorithm}. Reconstruction metric include Euclidean, $L_1$ distance or perceptual losses such as SSIM \cite{wang2004image}. The main weaknesses of reconstruction-based methods are i) difficulty of learning discriminative basis functions ii) finding effective similarity measures is non-trivial. Semi-supervised distribution-based approaches, attempt to learn the probability density function (PDF) of the normal data. Given a new sample, its probability is evaluated and is designated as anomalous if the probability is lower than a certain threshold. Such methods include: parametric models e.g. mixture of Gaussians (GMM). Non-parametric methods include Kernel Density Estimation \cite{latecki2007outlier} and kNN \cite{eskin2002geometric} (which we also consider reconstruction-based) The main weakness of distributional methods is the difficulty of density estimation for high-dimensional data. Another popular approach is one-class SVM \citep{scholkopf2000support} and related SVDD \cite{tax2004support}. SVDD can be seen as fitting the minimal volume sphere that includes at least a certain percentage of the normal data points. As this method is very sensitive to the feature space, kernel methods were used to learn an effective feature space.

\textit{Augmenting classical methods with deep networks:} The success of deep neural networks has prompted research combining deep learned features to classical methods. PCA methods were extended to deep auto-encoders \cite{yang2017towards}, while their reconstruction costs were extended to deep perceptual losses \cite{zhang2018unreasonable}. GANs were also used as a basis function for reconstruction in images. One approach \cite{zong2018deep} to improve distributional models is to first learn to embed data in a semantic, low dimensional space and then model its distribution using standard methods e.g. GMM. SVDD was extended by \citet{ruff2018deep} to learn deep features as a superior alternative for kernel methods. This method suffers from a "mode collapse" issue, which has been the subject of followup work. The approach investigated in this paper can be seen as belonging to this category, as classical kNN is extended with deep learned features.

\textit{Self-supervised Deep Methods:} Instead of using supervision for learning deep representations, self-supervised methods train neural networks to solve an auxiliary task for which obtaining data is free or at least very inexpensive. It should be noted that self-supervised representation typically under-perform those learned from large supervised datasets such as Imagenet. Auxiliary tasks for learning high-quality image features include: video frame prediction \citep{mathieu2015deep}, image colorization \citep{zhang2016colorful, larsson2016learning}, and puzzle solving \citep{noroozi2016unsupervised}. Recently, \citet{gidaris2018unsupervised} used a set of image processing transformations (rotation by $0,90,180,270$ degrees around the image axis), and predicted the true image orientation. They used it to learn high-quality image features. \citet{golan2018deep}, have used similar image-processing task prediction for detecting anomalies in images. This method has shown good performance on detecting images from anomalous classes. The performance of this method was improved by \citet{hendrycks2019using}, while it was combined with openset classification and extended to tabular data by \citet{bergman2020classification}. In this work, we show that self-supervised methods underperform simpler kNN-based methods that use strong generic feature extractors on image anomaly detection tasks.  

\section{Deep Nearest-Neighbors for Image Anomaly Detection}
\label{sec:method}

We investigate a simple K nearest-neighbors (kNN) based method for image anomaly detection. We denote this method, Deep Nearest-Neighbors (DN2).

\subsection{Semi-supervised Anomaly Detection}
\label{subsec:semi-supervised}

DN2 takes a set of input images $X_{train}=x_1,x_2..x_N$. In the semi-supervised setting we assume that all input images are normal. DN2 uses a pre-trained feature extractor $F$ to extract features from the entire training set:
\begin{equation}
    \label{eq:extract}
    f_i = F(x_i)
\end{equation}

In this paper, we use a ResNet feature extractor that was pretrained on the Imagenet dataset. At first sight it might appear that this supervision is a strong requirement, however such feature extractors are widely available. We will later show experimentally that the normal or anomalous images do not need to be particularly closely related to Imagenet. 

The training set is now summarized as a set of embeddings $F_{train} = f_1,f_2..f_N$. After the initial stage, the embeddings can be stored, amortizing the inference of the training set.

To infer if a new sample $y$ is anomalous, we first extract its feature embedding: $f_y = F(y)$. We then compute its kNN distance and use it as the anomaly score:
\begin{equation}
    \label{eq:knn}
    d(y) = \frac{1}{k} \sum_{f \in N_k(f_y)}{\|f - f_y\|^2}
\end{equation}
$N_k(f_y)$ denotes the $k$ nearest embeddings to $f_y$ in the training set  $F_{train}$. We elected to use the euclidean distance, which often achieves strong results on features extracted by deep networks, but other distance measures can be used in a similar way. By verifying if the distance $d(y)$ is larger than a threshold, we determine if an image $y$ is normal or anomalous. 

\subsection{Unsupervised Anomaly Detection}
\label{subsec:unsupervised}

In the fully-unsupervised case, we can no longer assume that all input images are normal, instead, we assume that only a small proportion of input images are anomalous. To deal with this more difficult setting (and inline with previous works on unsupervised anomaly detection), we propose to first conduct a cleaning
stage on the input images. After the feature extraction stage, we compute the kNN distance between each input image and the rest of the input images. Assuming that anomalous images lie in low density regions, we remove a fraction of the images with the largest kNN distances. This fraction should be chosen such that it is larger than the estimated proportion of anomalous input images. It will be later shown in our experiments that DN2 requires very few training images. We can therefore be very aggressive in the percentage of removed image, and keep only the images most likely to be normal (in practice we remove $50\%$ of training images). After removal of the suspected anomalous input images, the images are now assumed to have a very high-proportion of normal images. We can therefore proceed exactly as in the semi-supervised case.

\subsection{Group Image Anomaly Detection}
\label{subsec:group}

Group anomaly detection tackles the setting where the input sample consists of a set of images. The particular combination is important, but not the order. It is possible that each image in the set will individually be normal but the set as a whole will be anomalous. As an example, let us assume normal sets consisting of $M$ images, a randomly sampled image from each class. If we trained a point (per-image) anomaly detector, it will be able to detect anomalous sets containing pointwise anomalous images e.g. images taken from classes not seen in training. An anomalous set containing multiple images from one seen class, and no images from another will however be classified as normal as all images are individually normal. Previously, several deep autoencoder methods were proposed (e.g. \citet{dgroup}) to tackle group anomaly detection in images. Such methods suffer from multiple drawbacks: i) high sample complexity ii) sensitivity to reconstruction metric iii) potential lack of sensitivity to the groups. We propose an effective kNN based approach. The proposed method embeds the set by orderless-pooling (we chose averaging) over all the features of the images in the set: 
\begin{table*}[ht]
  \centering
  \caption{Anomaly Detection Accuracy on Cifar10 (ROCAUC $\%$)}
  \label{tab:exp_cifar10}

    \begin{tabular}{lcccccc}
    \toprule      

   & OC-SVM & Deep SVDD & GEOM & GOAD & MHRot & DN2 \\
    \midrule
   0 & 70.6 & 61.7 $\pm$ 1.3 & 74.7 $\pm$ 0.4 & 77.2 $\pm$ 0.6 & 77.5& \textbf{93.9}\\
   1 & 51.3 & 65.9 $\pm$ 0.7 & 95.7 $\pm$ 0.0 & 96.7 $\pm$ 0.2 & 96.9& \textbf{97.7}\\
   2 & 69.1 & 50.8 $\pm$ 0.3 & 78.1 $\pm$ 0.4 & 83.3 $\pm$ 1.4 & \textbf{87.3}& 85.5\\
   3 & 52.4 & 59.1 $\pm$ 0.4 & 72.4 $\pm$ 0.5 & 77.7 $\pm$ 0.7 & 80.9& \textbf{85.5}\\
   4 & 77.3 & 60.9 $\pm$ 0.3 & 87.8 $\pm$ 0.2 & 87.8 $\pm$ 0.7 & 92.7& \textbf{93.6}\\
   5 & 51.2 & 65.7 $\pm$ 0.8 & 87.8 $\pm$ 0.1 & 87.8 $\pm$ 0.6 & 90.2& \textbf{91.3}\\
   6 & 74.1 & 67.7 $\pm$ 0.8 & 83.4 $\pm$ 0.5 & 90.0 $\pm$ 0.6 & 90.9& \textbf{94.3}\\
   7 & 52.6 & 67.3 $\pm$ 0.3 & 95.5 $\pm$ 0.1 & 96.1 $\pm$ 0.3 & \textbf{96.5}& 93.6\\
   8 & 70.9 & 75.9 $\pm$ 0.4 & 93.3 $\pm$ 0.0 & 93.8 $\pm$ 0.9 & 95.2& \textbf{95.1}\\
   9 & 50.6 & 73.1 $\pm$ 0.4 & 91.3 $\pm$ 0.1 & 92.0 $\pm$ 0.6 & 93.3& \textbf{95.3}\\
   \midrule
   Avg & 62.0 & 64.8 & 86.0 & 88.2 & 90.1 & \textbf{92.5}\\
	 \bottomrule
    \end{tabular}
\end{table*}

\begin{enumerate}
    \item Feature extraction from all images in the group $g$, \\ $f^i_g = F(x^i_g)$
    \item Orderless pooling of features across the group: \\ $f_g = \frac{\sum_i f^i_g}{number~of~images}$
\end{enumerate}

Having extracted the group feature described above we proceed to detect anomalies using DN2.

\section{Experiments}
\label{sec:exp}

In this section, we present extensive experiments showing that the simple kNN approach described above achieves better than state-of-the-art performance. The conclusions generalize across tasks and datasets. We extend this method to be more robust to noise, making it applicable to the unsupervised setting. We further extend this method to be effective for group anomaly detection.

\subsection{Unimodal Anomaly Detection}
\label{subsec:exp:uni}

The most common setting for evaluating anomaly detection methods is unimodal. In this setting, a classification dataset is adapted by designating one class as normal, while the other classes as anomalies. The normal training set is used to train the method, all the test data are used to evaluate the inference performance of the method. In line with previous works, we report the ROC area under the curve (ROCAUC).

\begin{table}
  \centering
  \caption{Anomaly Detection Accuracy on Fashion MNIST and CIFAR10 (ROCAUC $\%$)}

    \begin{tabular}{lcccccc}
    \toprule      

   & OC-SVM & GEOM & GOAD & DN2 \\
    \midrule
   FashionMNIST & 92.8 & 93.5 & 94.1  & \textbf{94.4}\\
   CIFAR100 &  62.6 & 78.7 & - & \textbf{89.3} \\

	 \bottomrule
    \end{tabular}
    \label{tab:exp_small_extra}
\end{table}

We conduct experiments against state-of-the-art methods, deep-SVDD \cite{ruff2018deep} which combines OCSVM with deep feature learning. Geometric \cite{golan2018deep}, GOAD \cite{bergman2020classification}, Multi-Head RotNet (MHRot) \cite{hendrycks2019using}. The latter three all use variations of RotNet. 

For all methods except DN2, we reported the results from the original papers if available. In the case of Geometric \cite{golan2018deep} and the multi-head RotNet (MHRot) \cite{hendrycks2019using}, for datasets that were not reported by the authors, we run the Geometric code-release for low-resolution experiments, and MHRot for high-resolution experiments (as no code was released for the low-resolution experiments).   

\textit{Cifar10:} This is the most common dataset for evaluating unimodal anomaly detection. CIFAR10 contains $32 \times 32$ color images from 10 object classes. Each class has $5000$ training images and $1000$ test images. The results are presented in Tab.~\ref{tab:exp_cifar10}, note that the performance of DN2 is deterministic for a given train and test set (no variation between runs). We can observe that OC-SVM and Deep-SVDD are the weakest performers. This is because both the raw pixels as well as features learned by Deep-SVDD are not discriminative enough for the distance to the center of the normal distribution to be successful. Geometric and later approaches GOAD and MHRot perform fairly well but do not exceed $90\%$ ROCAUC. DN2 significantly outperforms all other methods.

In this paper, we choose to evaluate the performance of without finetuning between the dataset and simulated anomalies (which improves performance on all methods including DN2). Outlier Exposure is one technique for such finetuning. Although it does not achieve the top performance by itself, it reported improvements when combined with MHRot to achieve an average ROCAUC of $95.8\%$ on CIFAR10. This and other ensembling methods can also improve the performance of DN2 but are out-of-scope of this paper.    

\textit{Fashion MNIST:} We evaluate Geometric, GOAD and DN2 on the Fashion MNIST dataset consisting of 6000 training images per class and a test set of 1000 images per class.  We present a comparison of DN2 vs. OCSVM, Deep SVDD, Geometric and GOAD. We can see that DN2 outperforms all other methods, despite the data being visually quite different from Imagenet from which the features were extracted.

\textit{CIFAR100:} We evaluate Geometric, GOAD and DN2 on the CIFAR100 dataset. CIFAR100 has 100 fine-grained classes with 500 train images each or 20 coarse-grained classes with 2500 train images each. Following previous papers, we use the coarse-grained version. The protocol is the same as CIFAR10. We present a comparison of DN2 vs. OCSVM, Deep SVDD, Geometric and GOAD. The results are inline with those obtained for CIFAR10.

\textbf{Comparisons against MHRot:}

We present a further comparison between DN2 and MHRot \cite{hendrycks2019using} on several commonly-used datasets. The experiments give further evidence for the generality of DN2, in datasets where RotNet-based methods are not restricted by low-resolution, or by image invariance to rotations.

We compute the ROCAUC score on each of the first $20$ categories (all categories if there are less than $20$),  by alphabetical order, designated as normal for training. The standard train and test splits are used. All test images from all classes are used for inference, with the appropriate class designated normal and all the rest as anomalies. For brevity of presentation, the average ROCAUC score of the tested classes is reported.

\textit{$102$ Category Flowers \cite{nilsback2008automated}:} This dataset consists of $102$ categories of flowers, consisting of $10$ training images each. The test set consists of $30$ to over $200$ images per-class. 

\textit{Caltech-UCSD Birds $200$ \cite{wah2011caltech}:} This dataset consists of $200$ categories of bird species. Classes typically contain between $55$ to $60$ images split evenly between train and test. 

\textit{ CatsVsDogs \cite{elson2007asirra}:} This dataset consists of $2$ categories; dogs and cats with $10,000$ training images each. The test set consist of $2,500$ images for each class. Each image contains either a dog or a cat in various scenes and taken from different angles. The data was extracted from the ASIRRA dataset, we split each class to the first $10,000$ images as train and the last $2,500$ as test.

The results are shown in Tab.~\ref{tab:exp_mhrot}. DN2 significantly outperforms MHRot on all datasets.

\begin{table}
  \centering
  \caption{MHRot vs. DN2 on Flowers, Birds, CatsVsDogs (Average Class ROCAUC $\%$)}
  \label{tab:exp_mhrot}

    \begin{tabular}{lcc}
    \toprule      

 Dataset  &  MHRot & DN2 \\
    \midrule
   Oxford Flowers & 65.9 & \textbf{93.9}\\
   UCSD Birds 200 & 64.4 & \textbf{95.2}\\
   CatsVsDogs & 88.5 & \textbf{97.5}\\
	 \bottomrule
    \end{tabular}
\end{table}

\begin{figure}
  \centering

    \begin{tabular}{ccc}

   \includegraphics[scale=0.14]{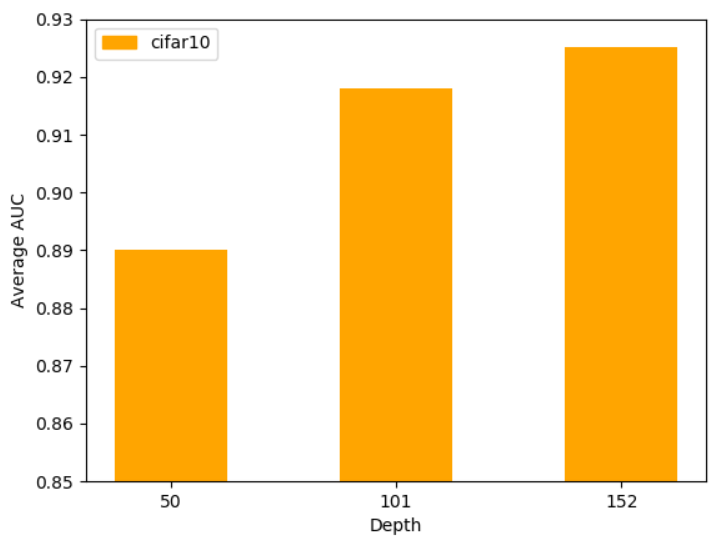} &
   \includegraphics[scale=0.14]{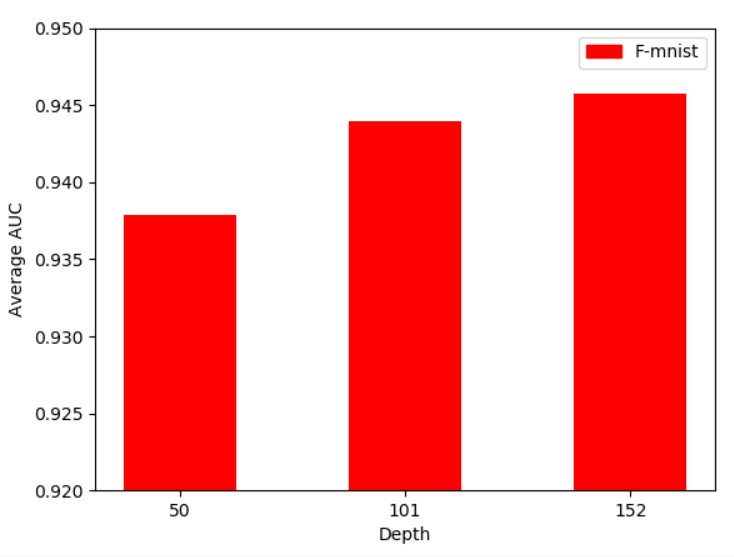} \\
    \end{tabular}
    \caption{Network depth (number of ResNet layers) improves both Cifar10 and FashionMNIST results.}
    \label{fig:small_depth}
\end{figure}

\begin{figure}
  \centering

    \begin{tabular}{c}

   \includegraphics[scale=0.2]{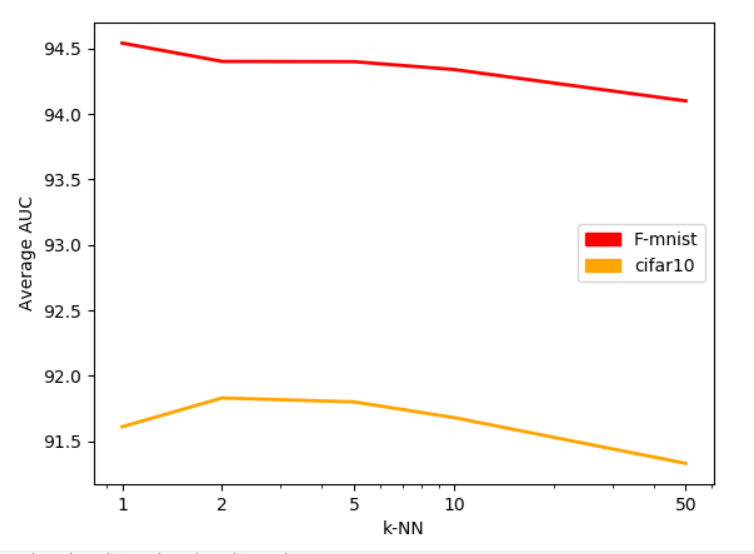} \\
    \end{tabular}
    \caption{Number of neighbors vs ROCAUC, the optimal number of K is around $2$.}
     \label{fig:small_neighbors}
\end{figure}

\textbf{Effect of network depth:}

Deeper networks trained on large datasets such as Imagenet learn features that generalize better than shallow network. We investigated the performance of DN2 when using features from networks of different depths. Specifically, we plot the average ROCAUC for ResNet with 50, 101, 152 layers in Fig.~\ref{fig:small_depth}. DN2 works well with all networks but performance is improved with greater network depth. 

\textbf{Effect of the number of neighbors:}

The only free parameter in DN2 is the number of neighbors used in kNN. We present in Fig.~\ref{fig:small_neighbors}, a comparison of average CIFAR10 and FashionMNIST ROCAUC for different numbers of nearest neighbors. The differences are not particularly large, but $2$ neighbors are usually best.

\textbf{Effect of data invariance:}

\begin{figure}
  \centering
  
  \label{fig:wbc_dior_samples}

    \begin{tabular}{cc}

   \includegraphics[scale=0.38]{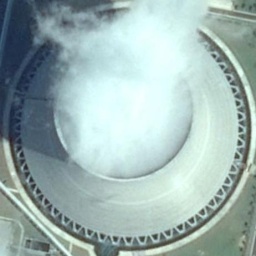} & \includegraphics[scale=0.8]{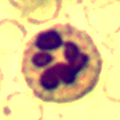}
    \end{tabular}
    \caption{(left) A chimney image from the DIOR dataset (right) An image from the WBC Dataset.}
\end{figure}

\begin{table}
  \centering
  \caption{Anomaly Detection Accuracy on DIOR and WBC (ROCAUC $\%$)}

    \begin{tabular}{lcclcc}
     \toprule      

      Dataset & MHRot & DN2 \\
      \midrule
      DIOR & 83.2 & \textbf{92.2} \\
      WBC & 60.5 & \textbf{82.9}\\
\bottomrule
    \end{tabular}
     \label{tab:exp_inv}
\end{table}

Methods that rely on predicting geometric transformations e.g. \cite{golan2018deep, hendrycks2019using, bergman2020classification}, use a strong data prior that images have a predetermined orientation (for rotation prediction) and centering (for translation prediction). This assumption is often false for real images. Two interesting cases not satisfying this assumption, are aerial and microscope images, as they do not have a preferred orientation, making rotation prediction ineffective.

\textit{DIOR \cite{li2020object}:} An aerial image dataset. The images are registered but do not have a preferred orientation. The dataset consists of $19$ object categories that have more than $50$ images with resolution above $120 \times 120$ (the median number of images per-class is $578$). We use the bounding boxes provided with the data, and take each object with a bounding box of at least $120$ pixels in each axis. We resize it to $256 \times 256$ pixels. We follow the same protocol as in the earlier datasets. As the images are of high-resolution, we use the public code release of Hendrycks \cite{hendrycks2018deep} as a self-supervised baseline. The results are summarized in Tab.~\ref{tab:exp_inv}. We can see that DN2 significantly outperforms MHRot. This is due both to the generally stronger performance of the feature extractor as well as the lack of rotational prior that is strongly used by RotNet-type methods. Note that the images are centered, a prior used by the MHRot translation heads.

\textit{WBC \cite{zheng2018fast}:} To further investigate the performance on difficult real world data, we performed an experiment on the WBC Image Dataset, which consists of high-resolution microscope images of different categories of white blood cells. The data do not have a preferred orientation. Additionally the dataset is very small, only a few tens of images per-class. We use Dataset $1$ that was obtained from Jiangxi Telecom Science Corporation, China, and split it to the $4$ different classes that contain more than $20$ images each. We set the first $80\%$ images in each class to the train set, and the last $20\%$ to the test set. The results are presented in Tab.~\ref{tab:exp_inv}. As expected, DN2 outperforms MHRot by a significant margin showing its greater applicability to real world data. 

\subsection{Multimodal Anomaly Detection}
\label{subsec:exp:multi}

\begin{table}
  \centering
  \caption{Anomaly Detection Accuracy on Multimodal Normal Image Distributions (ROCAUC $\%$)}

    \begin{tabular}{lcc}
     \toprule      

      Dataset & Geometric & DN2 \\
      \midrule
      CIFAR10 & 61.7 & \textbf{71.7}   \\
      CIFAR100 & 57.3 & \textbf{71.0} \\
      \bottomrule
    \end{tabular}
    \label{tab:exp_multi}
\end{table}

It has been argued (e.g. \citet{ahmed2019detecting}) that unimodal anomaly detection is less realistic as in practice, normal distributions contain multiple classes. While we believe that both settings occur in practice, we also present results on the scenario where all classes are designated as normal apart from a single class that is taken as anomalous (e.g. all CIFAR10 classes are normal apart from "Cat"). Note that we do not provide the class labels of the different classes that compose the normal class, rather we consider them to be a single multimodal class. We believe this simulates the realistic case of having a complex normal class consisting of many different unlabelled types of data.

We compared DN2 against Geometric on CIFAR10 and CIFAR100 on this setting. We provide the average ROCAUC across all the classes in Tab.~\ref{tab:exp_multi}. DN2 achieves significantly stronger performance than Geometric. We believe this is occurs as Geometric requires the network not to generalize on the anomalous data. However, once the training data is sufficiently varied the network can generalize even on unseen classes, making the method less effective. This is particularly evident on CIFAR100.  

\begin{figure}
  \centering
  
  \label{fig:group}

   \includegraphics[scale=0.4]{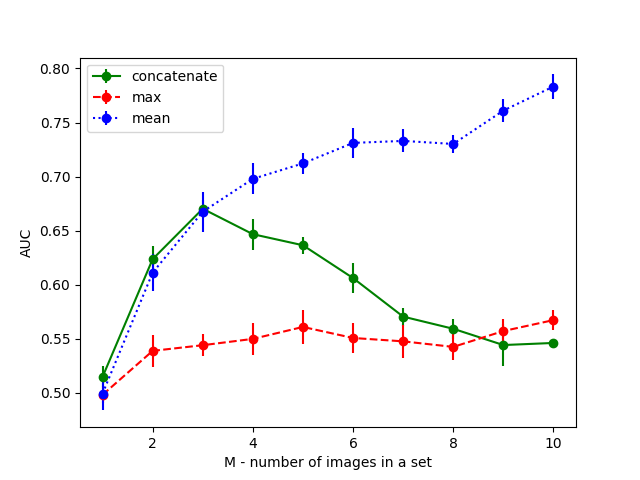}
    \caption{Number of images per group vs. detection ROCAUC. Group anomaly detection with mean pooling is better than simple feature concatenation for groups with more than $3$ images.}
\end{figure}

\begin{figure*}
  \centering

    \begin{tabular}{ccc}

   \includegraphics[scale=0.2]{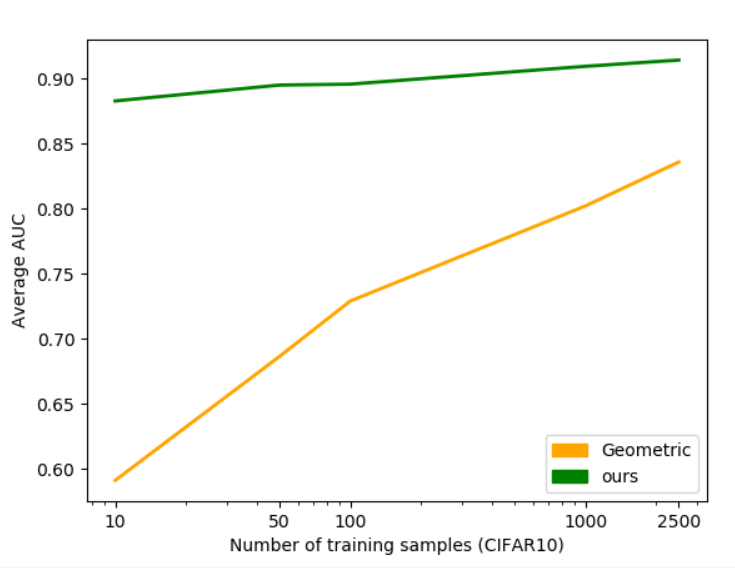} & \includegraphics[scale=0.2]{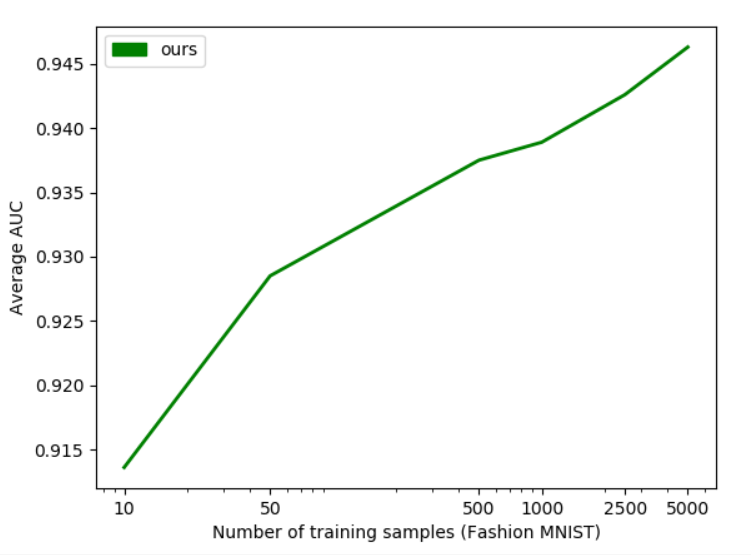} & \includegraphics[scale=0.2]{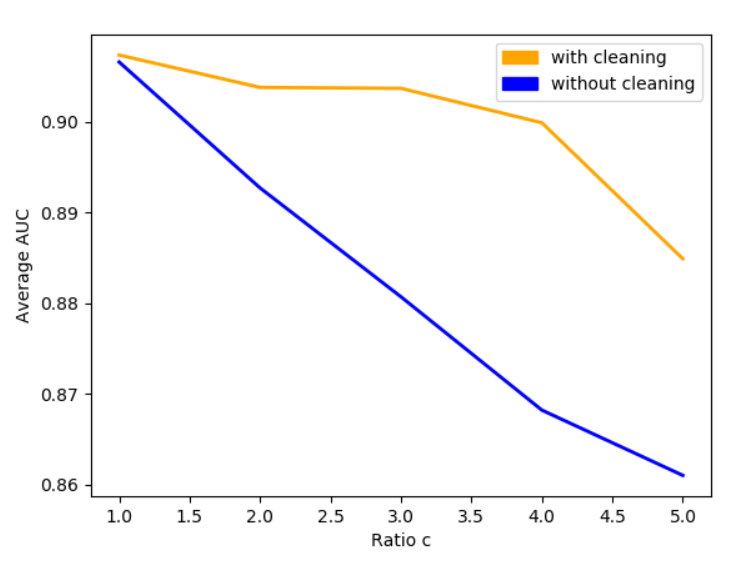} 
    \end{tabular}
    \caption{Number of training images vs. ROCAUC (left) CIFAR10 - Strong perfromance is achieved by DN2 even from 10 images, whereas Geometric deteriorates critically. (center) FashionMNIST - similarly strong performance by DN2. (right) Impurity ratio vs ROCAUC on CIFAR10. The training set cleaning procedure, significantly improves performance.}
    \label{fig:small_unsupervised}
\end{figure*}

\subsection{Generalization from Small Training Datasets}
\label{subsec:exp:small}

One of the advantage of DN2, which does not utilize learning on the normal dataset is its ability to generalize from very small datasets. This is not possible with self-supervised learning-based methods, which do not learn general enough features to generalize to normal test images. A comparison between DN2 and Geometric on CIFAR10 is presented in Fig.~\ref{fig:small_unsupervised}. We plotted the number of training images vs. average ROCAUC. We can see that DN2 can detect anomalies very accurately even from 10 images, while Geometric deteriorates quickly with decreasing number of training images. We also present a similar plot for FashionMNIST in Fig.~\ref{fig:small_unsupervised}. Geometric is not shown as it suffered from numerical issues for small numbers of images. DN2 again achieved strong performance from very few images.

\subsection{Unsupervised Anomaly Detection}
\label{subsec:exp:unsupervised}

There are settings where the training set does not consist of purely normal images, but rather a mixture of unlabelled normal and anomalous images. Instead we assume that anomalous images are only a small fraction of the number of the normal images. The performance of DN2 as function of the percentage of anomalies in the training set is presented in Fig.~\ref{fig:small_unsupervised}. The performance is somewhat degraded as the percentage of training set impurities exist. To improve the performance, we proposed a cleaning stage, which removes $50\%$ of the training set images that have the most distant $kNN$ inside the training set. We then run DN2 as usual. The performance is also presented in Fig.~\ref{fig:small_unsupervised}. Our cleaning procedure is clearly shown to significantly improve the performance degradation as percentage of impurities.

\subsection{Group Anomaly Detection}
\label{subsec:exp:group}

To compare to existing baselines, we first tested our method on the task in \citet{dgroup}. The data consists of normal sets containing $10-50$ MNIST images of the same digit, and anomalous sets containing $10-50$ images of different digits. By simply computing the trace-diagonal of the covariance matrix of the per-image ResNet features in each set of images, we achieved $0.92$ ROCAUC vs. $0.81$ in the previous paper (without using the training set at all). 

As a harder task for group anomaly detection in unordered image sets, we designate the normal class as sets consisting of exactly one image from each of the $M$ CIFAR10 classes (specifically the classes with ID $0..M-1$) while each anomalous set consisted of $M$ images selected randomly among the same classes (some classes had more than one image and some had zero). As a simple baseline, we report the average ROCAUC (Fig,~\ref{fig:group}) for anomaly detection using DN2 on the concatenated features of each individual image in the set. As expected, this baseline works well for small values of $M$ where we have enough examples of all possible permutations of the class ordering, but as $M$ grows larger ($M>3$), its performance decreases, as the number permutations grows exponentially. We compare this method, with 1000 image sets for training, to nearest neighbours of the orderless max-pooled and average-pooled features, and see that mean-pooling significantly outperforms the baseline for large values of $M$. While we may improve the performance of the concatenated features by augmenting the dataset with all possible orderings of the training sets, it is will grow exponentially for a non-trivial number of $M$ making it an ineffective approach.

\subsection{Implementation}
\label{subsec:exp:imp}

In all instances of DN2, we first resize the input image to $256 \times 256$, we take the center crop of size $224 \times 224$, and using an Imagenet pre-trained ResNet ($101$ layers unless otherwise specified) extract the features just after the global pooling layer. This feature is the image embedding.

\section{Analysis}
\label{sec:analysis}

In this section, we perform an analysis of DN2, both by comparing kNN to other classification methods, as well as comparing the features extracted by the pretrained networks vs. features learned by self-supervised methods. 

\subsection{kNN vs. one-class classification}
\label{subsec:analysis:knn}

\begin{figure*}[ht]
  \centering

    \begin{tabular}{ccc}

   \includegraphics[scale=0.4, trim=0 0 100 0 , clip]{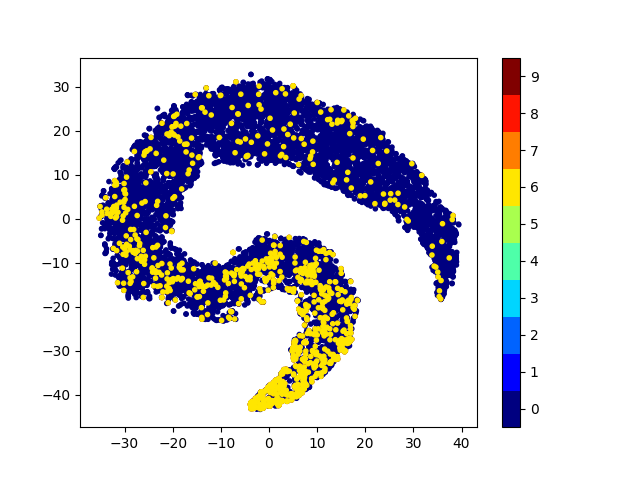} &
   \includegraphics[scale=0.4, trim=0 0 100 0 , clip]{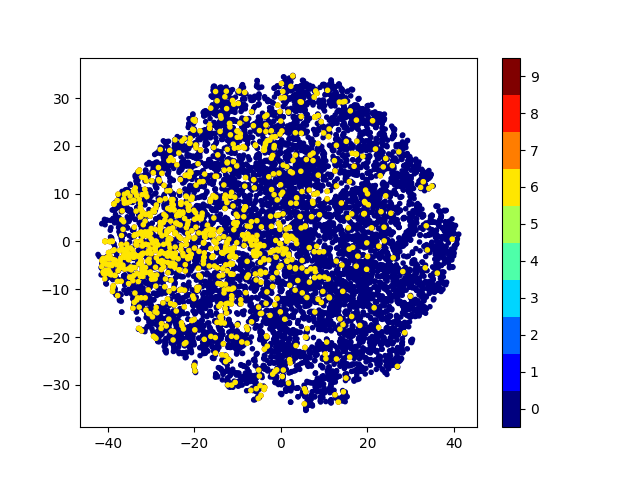} &
   \includegraphics[scale=0.4, trim=0 0 100 0 , clip]{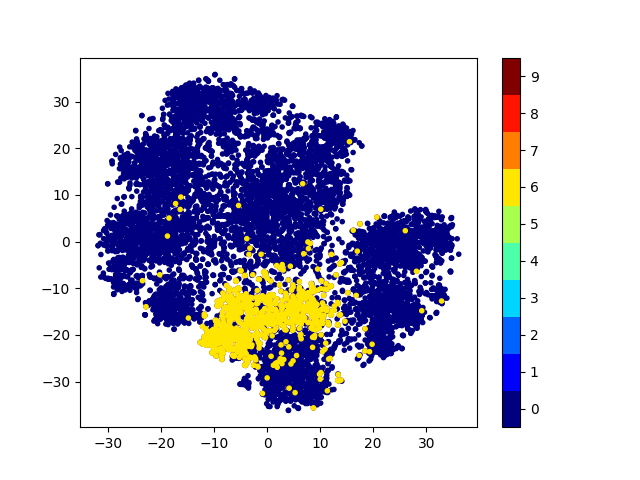} \\
   \includegraphics[scale=0.4, trim=0 0 100 0 , clip]{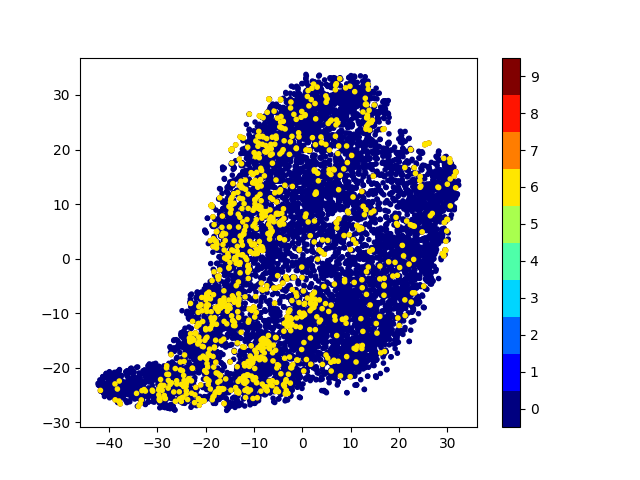} &
   \includegraphics[scale=0.4, trim=0 0 100 0 , clip]{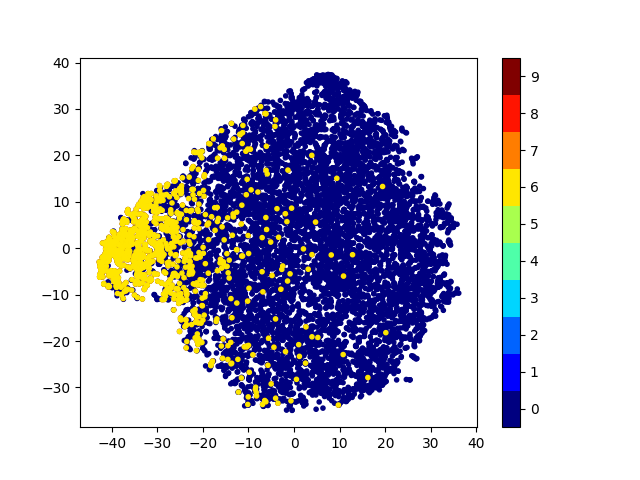} &
   \includegraphics[scale=0.4, trim=0 0 100 0 , clip]{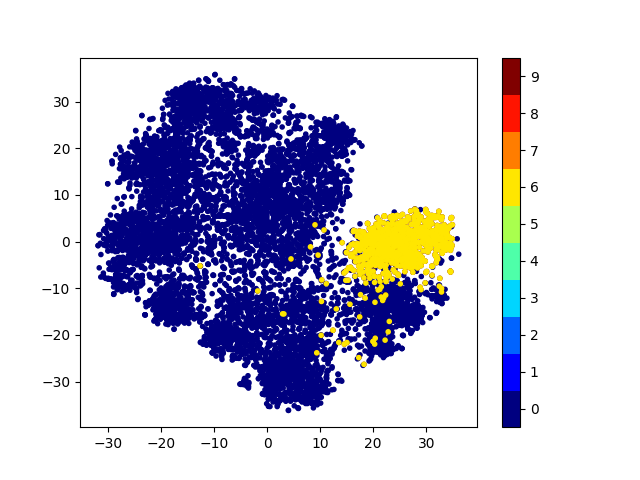} \\
    \end{tabular}
    \caption{t-SNE plots of the features learned by SVDD (left), Geometric (center) and Imagenet pre-trained (right) on CIFAR10, where the normal class is Airplane (top), Automobile (bottom). We can see that Imagenet-pretrained features clearly separate the normal class (yellow) and anomalies (blue). Geometric learns poor features of Airplane and reasonable features on Automobile. Deep-SVDD does not learn features that allow clean separation.  }
    \label{fig:tsne}
\end{figure*}

In our experiments, we found that kNN achieved very strong performance for anomaly detection tasks. Let us try to gain a better understanding of the reasons for the strong performance. In Fig.~\ref{fig:tsne} we can observe t-SNE plots of the test set features of CIFAR10. The normal class is colored in yellow while the anomlous data is marked in blue. It is clear that the pre-trained features embed images from the same class into a fairly compact region. We therefore expect the density of normal training images to be much higher around normal test images than around anomalous test images. This is responsible for the success of kNN methods.

\begin{table}
  \centering
  \caption{Accuracy on CIFAR10 using K-means approximations and full kNN (ROCAUC $\%$)}
  \label{tab:exp_kmeans}

    \begin{tabular}{ccccc}
     \toprule      

      C=1 & C=3 & C=5 & C=10 & kNN \\
      \midrule
      91.94& 92.00& 91.87& 91.64& 92.52 \\
      \bottomrule

    \end{tabular}
\end{table}

kNN has linear complexity in the number of training data samples. Methods such as One-Class SVM or SVDD attempt to learn a single hypersphere, and use the distance to the center of the hypersphere as a measure of anomaly. In this case the inference runtime is constant in the size of the training set, rather than linear as in the kNN case. The drawback is the typical lower performance. Another popular way \cite{fukunaga1975branch} of decreasing the inference time is using K-means clustering of the training features. This speeds up inference by a ratio of $\frac{N}{K}$. We therefore suggest speeding up DN2 by clustering the training features into $K$ clusters and the performing kNN on the clusters rather than the original features. Tab.~\ref{tab:exp_kmeans} presents a comparison of performance of DN2 and its K-means approximations with different numbers of means (we use the sum of the distances to the 2 nearest neighbors). We can see that for a small loss in accuracy, the retrieval speed can be reduced significantly. 

\subsection{Pretrained vs. self-supervised features}
\label{subsec:analysis:features}

To understand the improvement in performance by pretrained feature extractors, we provide t-SNE plots of normal and anomalous test features extracted by Deep-SVDD, Geometric and DN2 (Resnet50 pretrained on Imagenet). The top plots are of a normal class that achieves moderate detection accuracy, while the bottom plots are of a normal class that achieves high accuracy. We can immediately observe that the normal class in Deep-SVDD is scattered among the anomalous classes, explaining its lower performance. In Geometric the features of the normal class are a little more localized, however the density of the normal region is still only moderately concentrated. We believe that the fairly good performance of Geometric is achieved by the massive ensembling that it performs (combination of $72$ augmentations). We can see that Imagenet pretrained features preserve very strong locality. This explains the strong performance of DN2. 

\section{Discussion}
\label{sec:disc}

\textbf{A general paradigm for anomaly detection:} Recent papers (e.g. \citet{golan2018deep}) advocated the paradigm of self-supervision, possibly with augmentation by an external dataset e.g. outlier exposure. The results in this paper, give strong evidence to an alternative paradigm: i) learn general features using all the available supervision on vaguely related datasets ii ) the learned features are expected to be general enough to be able to use standard anomaly detection methods (e.g. kNN, k-means). The pretrained paradigm is much faster to deploy than self-supervised methods and has many other advantages investigated extensively in Sec.~\ref{sec:exp}. We expect that for image data that has no similarity whatsoever to Imagenet, using pre-trained features may be less effective. That withstanding, in our experiments, we found that Imagenet-pretrained features were effective on aerial images as well as microscope images, while both settings are very different from Imagenet. We therefore expect DN2-like methods to be very broadly applicable.

\textbf{External supervision:} The key enabler for DN2's success is the availability of a high quality external feature extractor. The ResNet extractor that we used was previously trained on Imagenet. Using supervision is typically seen as being more expensive and laborious than self-supervised methods. In this case however, we do not see it as a disadvantage at all. We used networks that have already been trained and are as commoditized as free open-source software libraries. They are available completely free, no new supervision at all is required for using such networks for any new dataset, as well as minimal time or storage costs for training. The whole process consists of merely a single PyTorch line, we therefore believe that in this case, the discussion of whether these methods can be considered supervised is purely philosophical.

\textbf{Scaling up to very large datasets:} Nearest neighbors are famously slow for large datasets, as the runtime increases linearly with the amount of training data. The complexity is less severe for parametric classifiers such as neural networks. As this is a well known issue with nearest neighbors classification, much work was performed at circumventing it. One solution is fast kNN retrieval e.g. by kd-trees. Another solution used in Sec.~\ref{sec:analysis}, proposed to speed up kNN by reducing the training set through computing its k-means and computing kNN on them. This is generalized further by an established technique that approximates NN by a recursive K-means algorithm \cite{fukunaga1975branch}. We expect that in practice, most of the runtime will be a result of the neural network inference on the test image, rather than on nearest neighbor retrieval. 

\textbf{Non-image data:} Our investigation established a very strong baseline for image anomaly detection. This result, however, does not necessarily mean that all anomaly detection tasks can be performed this way. Generic feature extractors are very successful on images, and are emerging in other tasks e.g. natural language processing (BERT \cite{devlin2018bert}). This is however not the case in some of the most important areas for anomaly detection i.e. tabular data and time series. In those cases, general feature extractors do not exist, and due to the very high variance between datasets, there is no obvious path towards creating such feature extractors. Note however that as deep methods are generally less successful on tabular data, the baseline of kNN on raw data is a very strong one. That withstanding, we believe that these data modalities present the most promising area for self-supervised anomaly detection. \citet{bergman2020classification} proposed a method along these lines.

\section{Conclusion}
\label{sec:conc}

We compare a simple method, kNN on deep image features, to current approaches for semi-supervised and unsupervised anomaly detection. Despite its simplicity, the simple method was shown to outperform the state-of-the-art methods in terms of accuracy, training time, robustness to input impurities, robustness to dataset type and sample complexity. Although, we believe that more complex approaches will eventually outperform this simple approach, we think that DN2 is an excellent starting point for practitioners of anomaly detection as well as an important baseline for future research. 

\bibliography{example_paper}
\bibliographystyle{icml2020}

\end{document}